\documentclass[journal,twoside]{IEEEtran}

\pdfoutput=1
\usepackage{algorithm}
\usepackage{algorithmic}
\usepackage{multirow}
\usepackage{float}
\usepackage{array}
\usepackage{graphicx}
\usepackage{subfigure}
\usepackage{booktabs}
\graphicspath{{figure/}{figures/}}

\usepackage{amsthm}
\usepackage{amsmath}
\usepackage{amssymb}
\usepackage{color}

\usepackage{fancyhdr}
\usepackage{tikz,xcolor,hyperref}
 
\definecolor{lime}{HTML}{A6CE39}
\DeclareRobustCommand{\orcidicon}{%
	\begin{tikzpicture}
	\draw[lime, fill=lime] (0,0) 
	circle [radius=0.16] 
	node[white] {{\fontfamily{qag}\selectfont \tiny ID}};
	\draw[white, fill=white] (-0.0625,0.095) 
	circle [radius=0.007];
	\end{tikzpicture}
	\hspace{-2mm}
}
\hypersetup{hidelinks} 
\foreach \x in {A, ..., Z}{%
	\expandafter\xdef\csname orcid\x\endcsname{\noexpand\href{https://orcid.org/\csname orcidauthor\x\endcsname}{\noexpand\orcidicon}}
}
 
\begin{document}
%
% paper title
% Titles are generally capitalized except for words such as a, an, and, as,
% at, but, by, for, in, nor, of, on, or, the, to and up, which are usually
% not capitalized unless they are the first or last word of the title.
% Linebreaks \\ can be used within to get better formatting as desired.
% Do not put math or special symbols in the title.
\title{Viewpoint-aware Progressive Clustering for Unsupervised Vehicle Re-identification}

\author{Aihua~Zheng,~\IEEEmembership{}%Member,~IEEE,
        Xia Sun,~\IEEEmembership{}
        Chenglong Li,~\IEEEmembership{}
			Jin Tang~\IEEEmembership{}
\thanks{
This research is supported in part by the Major Project for New Generation of AI under Grant (No. 2018AAA0100400), National Natural Science Foundation of China (61976002, 61976003 and 61860206004), the Natural Science Foundation of Anhui Higher Education Institutions of China (KJ2019A0033), and the Open Project Program of the National Laboratory of Pattern Recognition (NLPR) (201900046). (Corresponding author: Chenglong Li.)
}
\thanks{A. Zheng, X. Sun, C. Li, and J. Tang are with the Anhui Provincial Key Laboratory of Multi-modal Cognitive Computation, School of Computer Science and Technology, Anhui University, Hefei, 230601, China
(e-mail: ahzheng214@foxmail.com; sunxia233@foxmail.com; lcl1314@foxmail.com;
tangjin@ahu.edu.cn)}
}

% <-this % stops a space
%\thanks{}% <-this % stops a space M. Shell was with the Department
%of Electrical and Computer Engineering, Georgia Institute of Technology, Atlanta,
%GA, 30332 USA e-mail: (see http://www.michaelshell.org/contact.html).
%\thanks{}% <-this % stops a spaceJ. Doe and J. Doe are with Anonymous University.Manuscript received April 19, 2005; revised August 26, 2015.
%\thanks{}}

% make the title area
\maketitle
\thispagestyle{fancy}            %更改plain状态，首页格式设为fancy
\fancyhead{}                     %清除以前的命令
\rhead{ \scriptsize 1}
\cfoot{This work has been submitted to the IEEE for possible publication. Copyright may be transferred without notice, after which this version may no longer be accessible.}                 %页脚左侧内容

\renewcommand{\headrulewidth}{0pt}      %把页眉线的宽度设为零，即去掉页眉线
\renewcommand{\footrulewidth}{0pt}
 
             %首页后的章节格式设置为空
% As a general rule, do not put math, special symbols or citations
% in the abstract or keywords.
\begin{abstract}
Vehicle re-identification (Re-ID) is an active task due to its importance in large-scale intelligent monitoring in smart cities.
Despite the rapid progress in recent years, most existing methods handle vehicle Re-ID task in a supervised manner, which is both time and labor-consuming and limits their application to real-life scenarios. 
Recently, unsupervised person Re-ID methods achieve impressive performance by exploring domain adaption or clustering-based techniques.
However, one cannot directly generalize these methods to vehicle Re-ID since vehicle images present huge appearance variations in different viewpoints. 
To handle this problem, we propose a novel viewpoint-aware clustering algorithm for unsupervised vehicle Re-ID.
In particular, we first divide the entire feature space into different subspaces according to the predicted viewpoints and then perform a progressive clustering to mine the accurate relationship among samples. 
%zheng-need refine
%
Comprehensive experiments against the state-of-the-art methods on two multi-viewpoint benchmark datasets VeRi and VeRi-Wild validate the promising performance of the proposed method in both with and without domain adaption scenarios while handling unsupervised vehicle Re-ID.
\end{abstract}

% Note that keywords are not normally used for peerreview papers.
\begin{IEEEkeywords}
Viewpoint-aware, Progressive Clustering, Vehicle Re-ID, Unsupervised Learning.
\end{IEEEkeywords}

% For peer review papers, you can put extra information on the cover
% page as needed:
% \ifCLASSOPTIONpeerreview
% \begin{center} \bfseries EDICS Category: 3-BBND \end{center}
% \fi
%
% For peerreview papers, this IEEEtran command inserts a page break and
% creates the second title. It will be ignored for other modes.
\IEEEpeerreviewmaketitle
\section{Introduction}

\begin{figure}[t]
  \centering
  \includegraphics[scale=0.7]{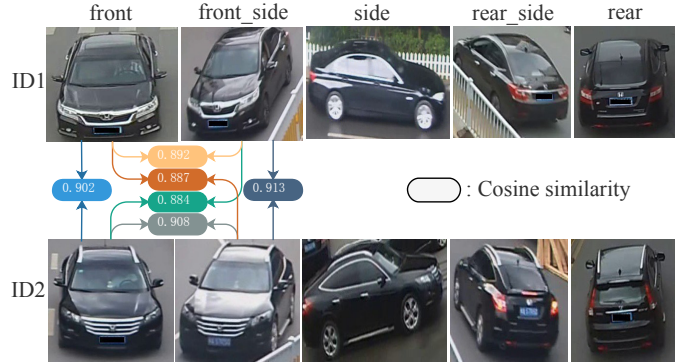}
  \caption{Illustration of major challenges in vehicle Re-ID. Different vehicles with the same viewpoint have higher visual similarity than those same vehicles with different viewpoints, and it is very common in real scenes that the same vehicle with different viewpoints sometimes has a similar appearance with those different vehicles with different viewpoints. These examples demonstrate that vehicle Re-ID greatly suffers from the challenges of large intra-class discrepancy and inter-class similarity.}
  \label{fig:sample1}
\vspace{-0.4cm}
\end{figure}

\IEEEPARstart{V}{ehicle} re-identification aims to identify a specific vehicle in non-overlapping camera networks.
It is a crucial task in modern society with potential applications in artificial transportation, smart city and public security, to name a few.
Similar to the person Re-ID task, vehicle Re-ID faces common challenges such as illumination and viewpoint changes across cameras, background clutters, and occlusions.
Besides, vehicle Re-ID dramatically suffers from the challenges of large intra-class discrepancy and inter-class similarity.
This is because different vehicles might present exactly similar appearance while the same vehicle might present totally different features, as shown in Fig.~\ref{fig:sample1}.
Therefore, one cannot directly deploy person Re-ID models to achieve satisfactory performance in vehicle Re-ID.

With the blossom of deep learning techniques and its powerful learning ability on large labeled data, various supervised learning architectures~\cite{liu2016deep,guo2018learning,zheng2019vehiclenet,he2019part,zhou2017cross, zhou2018vehicle, liu2018ram,wang2020discriminative,suprem2020looking} have been proposed and achieved remarkable performance for vehicle Re-ID.
Despite great progress, supervised learning-based methods require numerous annotations to train the deep models, which are time and labor-consuming and significantly limit real-life applications of vehicle Re-ID.

Domain adaptation, which transfers the learned information from the source domain (labeled data) to the target domain (unlabeled data), has been widely explored in the past decade as one of unsupervised learning manners in both person Re-ID~\cite{zhong2018generalizing,fan2018unsupervised,peng2016unsupervised,wang2018transferable} and vehicle Re-ID~\cite{peng2019cross,song2020unsupervised}.
However, they still require large annotations in the source domain. In addition, when the style gap between the two domains is too large, these transfer learning methods are also limited.
%
%Furthermore, unlike the common classification-based machine learning tasks transferring domain knowledge in close-set, the open-set issue, which means the source and target domains have different categories, brings an additional challenge for domain adaption Re-ID.
%

Different from domain adaptation-based methods, we study the problem of vehicle Re-ID in the target-only unsupervised learning framework, which does not rely on any labeled data in the source domain.
As one of the target-only unsupervised methods, clustering-based methods have been widely explored in the related computer vision tasks~\cite{caron2018deep,lin2019bottom,ding2019dispersion,zhang2019self,song2020unsupervised,Fu_2019_ICCV}.
Recent efforts on clustering-based methods in person Re-ID are to assign pseudo labels for samples by clustering algorithms and then use these labeled samples to train Re-ID models~\cite{lin2019bottom,ding2019dispersion,song2020unsupervised,zhao2020unsupervised}.

However, one can not directly apply these techniques to vehicle Re-ID.
%
%However, they mainly focus on the global comparison of the data while ignoring inter-instance similarity of different vehicle images in the same viewpoint, and large intra-instance discrepancy of the same vehicle in diverse viewpoints.
%
One of the key reasons is large viewpoint variations of vehicles, which bring big challenges to clustering algorithms. 
As shown in Fig.~\ref{fig:sample1}, by directly calculating the cosine similarity between vehicle images, we can see the similarity between the same vehicle images with different viewpoints is even lower than that between different vehicles in the same viewpoint with a similar appearance, which is referred as the similarity dilemma of vehicles in this paper.
Due to the inter-instance similarity and intra-instance discrepancy caused by large viewpoint variations of vehicles, the accuracy of clustering algorithms is significantly affected, and the performance of vehicle Re-ID would thus be extremely degraded.

To handle this problem, we propose a novel viewpoint-aware progressive clustering framework (VAPC) for robust unsupervised vehicle Re-ID.
We observe that vehicle images from different viewpoints of the same ID are more similar than vehicle images from different viewpoints of different IDs, e.g., image pairs \{ID1(\textit{front}), ID1(\textit{front\_side})\} are more similar than \{ID1(\textit{front}), ID2(\textit{front\_side})\}. 
Therefore we can divide the vehicles into different viewpoints. The vehicles in each viewpoint cluster exclude the effects of large viewpoint variations, and same ID from different viewpoints can be correctly classified according to the degree of similarity.
In addition, when clustering is performed only between samples of the same viewpoint, the comparison of different viewpoints with the same ID is excluded, which further reduces the intra-class differences and simplifies the clustering task.
Therefore, we propose a viewpoint-aware progressive clustering framework, which can be regarded as three parts.
First, considering the extreme viewpoint changes of the vehicle, we design a viewpoint-aware network, which can be pre-trained using viewpoint annotations~\cite{zheng2019attributes}, to predict viewpoints of vehicle images as the prior information. 
Second, feature extraction is crucial to the performance of clustering.
To extract the discriminative feature of each sample, it is necessary to train an initial model with strong feature extraction capabilities. 
In this paper, we use a self-supervised manner to learn the discriminative feature of each sample. 
Without the ground truth labels in the target-only unsupervised learning, we treat each sample as a category and force the network to learn the discriminative feature of each sample via the repelled loss~\cite{xiao2017joint,lin2019bottom}, which we call the recognition stage.
Third, we design a viewpoint-aware progressive clustering algorithm to handle the problem of similarity dilemma discussed above. 
Specifically, we first perform clustering in each vehicle image set with the same viewpoint and then cluster them by comparing the similarity of clusters across different viewpoints. 
In this way, we can distinguish small gaps between different identities in the same viewpoint, and mine the same identity samples with large gaps between different viewpoints.

We use the clustered results to train the Re-ID network in a supervised way after progressive viewpoint-aware clustering.
However, the clustering performance of different viewpoints significantly relies on the clustering results from the same viewpoint.
Therefore, we introduce the $k$-reciprocal encoding~\cite{zhong2017re,Fu_2019_ICCV,song2020unsupervised} as the distance metric to feature comparison of the same viewpoint due to its powerful ability in mining similar samples.

In addition, recent  methods~\cite{zhang2019self,song2020unsupervised,Fu_2019_ICCV} achieve remarkable performance on target-only unsupervised person Re-ID.
However, they directly employ the prevalent DBSCAN~\cite{ester1996density} to obtain pseudo labels while discarding all noisy samples (the hard positive and hard negative samples with pseudo labels assigned as -1) in the training stage. 
We argue that it is more important to learn the discriminative embeddings by mining hard positive samples than naively learning from simple samples, which has been proven in a large number of machine learning tasks~\cite{hermans2017defense,wang2020unsupervised, yu2019unsupervised,sohn2016improved,shi2016embedding,oh2016deep}.
To this end, we propose a noise selection method to classify each noise sample into a suitable cluster by the similarity between the noise sample and other clusters.

Based on the above discussion, VAPC focuses on addressing unsupervised vehicle Re-ID through a viewpoint-aware progressive clustering framework. 
We alleviate the impact of vehicle similarity dilemmas on clustering by transforming global comparisons into progressive clustering based on viewpoint.
To improve the clustering quality of the same viewpoint cluster, we introduced $k$-reciprocal encoding~\cite{zhong2017re,Fu_2019_ICCV,song2020unsupervised} as a distance metric for DBSCAN~\cite{ester1996density} clustering. 
In order to deal with outlier noise samples, we propose a noise selection method to improve the generalization ability of the model further.
The major contributions of this work are summarized as follows.
\begin{itemize}
  \item We propose a novel progressive clustering method to handle the similarity dilemma of vehicles in unsupervised vehicle Re-ID. To our best knowledge, this is the first time to employ the viewpoint-aware progressive clustering algorithm to achieve unsupervised vehicle Re-ID.
  %
%  \item We propose to embed the $k$-reciprocal encoding~\cite{zhong2017re,Fu_2019_ICCV,song2020unsupervised} in our framework to explore more samples of the same identity for robust clustering.
  %
  \item We designed a noise selection scheme to mine the hard positive samples with the same identity while considering their relationship to the hard negative samples, which significantly improves the discriminative ability of our network.
  \item Comprehensive experimental results on two benchmark datasets, including VeRi-776~\cite{liu2016deep} and VeRi-Wild~\cite{lou2019large} demonstrate the promising performance of our method and yield to a new state-of-the-art for unsupervised vehicle Re-ID.
  %
% You must have at least 2 lines in the paragraph with the drop letter
% (should never be an issue)
  %
% You must have at least 2 lines in the paragraph with the drop letter
% (should never be an issue)
\end{itemize}
%\hfill mds
%\hfill August 26, 2015
\section{Related Works}
Since most vehicle Re-ID methods are in a supervised fashion, we briefly review the progress in supervised vehicle Re-ID and recent advances in unsupervised person/vehicle Re-ID.
% needed in second column of first page if using \IEEEpubid
%\IEEEpubidadjcol

\subsection{Vehicle Re-ID.}
Most existing deep vehicle re-identification methods follow a supervised setting. 
Pioneer vehicle Re-ID methods~\cite{Zheng_2019_CVPR_Workshops,8653852,He_2019_CVPR} focus on the discriminative feature learning. Lou et al.~\cite{8653852} by mining similar negative samples, the features learned by the model are more robust. He et al.~\cite{He_2019_CVPR} proposed an efficient feature preserving method, which can enhance the perception ability of subtle differences.
%zheng-先总结一下先前的不考虑属性和角度等信息的车辆reid方法
%
Some works introduce~\cite{cormier2016low,liu2016large,liu2017provid,zhou2017cross,zhou2018vehicle,liu2018ram} additional attribute information, such as color or type to improve the discrimination of the deep feature for vehicle Re-ID. 
Temporal path information is also auxiliary information and has been widely employed~\cite{shen2017learning,wang2017orientation}, to improve the robustness of vehicle Re-ID, especially for the vehicles with a similar appearance from the same manufacture.
To handle the viewpoint variation issue in vehicle Re-ID, Zhou et al.~\cite{zhou2017cross,zhou2018vehicle} and Liu et al.~\cite{liu2018ram} employ GAN to infer multi-view information from a single-view of the input image in either image or feature level, to boost the performance by integrating the input and generated images or features.
Chu et al.~\cite{chu2019vehicle} separate the Re-ID into similar and different viewpoint modes, and learn the respective deep metric for each case. 
In the case of a known 3D bounding box for the vehicle image, Sochor et al.~\cite{sochor2016boxcars} calculated orientation information through 3D coordinates and added to the feature map to improve performance.
%zheng-focus on how to use 3D box.
%
%
Despite the significant progress on vehicle Re-ID, these supervised deep learning-based methods require extensive training data, which is expensive in both time and labor-consuming.
\begin{figure*}[!htbp]
  \centering
  \includegraphics[height=7cm,width=18.2cm]{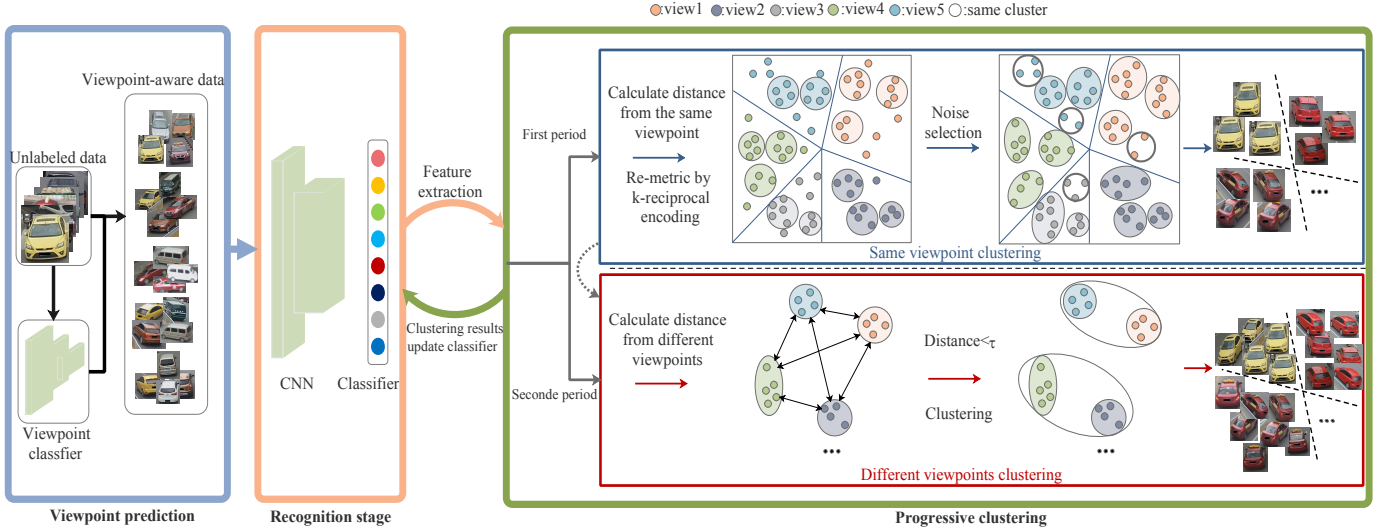}
  \caption{The overview of our method framework. We first predict each viewpoint, and then the viewpoint-aware unlabeled training set is input to the CNN model for feature extraction, which can be divided into different directional feature clusters. Then we will go through a recognition stage to make each sample feature extracted by the network more identifying. We design a clustering method that divides direction and period. In the first period, we cluster within the same viewpoint. For the noise samples found in the clustering process, we designed a noise selection method to select. In the second period, comparing the distances of all different viewpoints, clusters smaller than the distance threshold $\tau $ will be merged. The network is iteratively trained based on the final clustering results.}
  \label{fig:network}
\vspace{-0.3cm}
\end{figure*}

\subsection{Unsupervised Person/Vehicle Re-ID.}
Along with the great achievement on person Re-ID, unsupervised person Re-ID offers more challenges, which has attracted more and more attention recently. 
%zheng-update
%
Recent advances of unsupervised person Re-ID methods generally fall into
two categories. 1) The domain adaption based methods~\cite{zhong2018generalizing,fan2018unsupervised,peng2016unsupervised,wang2018transferable,deng2018image,wei2018person}, which aims to transfer the knowledge in the labeled source domain to the unlabeled target domain.
Although the domain adaption based methods make impressive achievement in unsupervised Re-ID by exploring domain-invariant features, they still require a large amount of label annotation in the source domain.
Furthermore, the huge diversity in different domains limits their transferring capabilities.
2) The target-only based methods~\cite{lin2019bottom,ding2019dispersion,lin2020unsupervised}, which fulfill the unsupervised task by dividing the unlabeled samples into different categories based on specific similarity.
Lin et al.~\cite{lin2019bottom} treat each image as a single category and then gradually reduces the number of categories in subsequent clusters.
Lin et al.~\cite{lin2020unsupervised} propose a framework that mines the similarity as a soft constraint and introduce camera information to encourage similar samples under different cameras to approach.
%zheng-talk sth about the challenge in clustering based methods.
%

To the best of our knowledge, there are few works on unsupervised vehicle Re-ID.
Peng et al.~\cite{peng2019cross} propose to use a style GAN to generate vehicle pictures in the source domain more like the target domain.
%zheng-generate what? source domain???not clear
%
They assume that the source domain contains more viewpoints than the target domain for a better generation.
Song et al.~\cite{song2020unsupervised} introduce the theoretical guarantees of unsupervised domain adaptive Re-ID based on and use a self-training scheme to iteratively optimize the unsupervised domain adaptation model. However, it only focuses on unsupervised domain adaptation, not target-only unsupervised learning.
Bashir et al.~\cite{bashir2019vr} employ clustering and reliable result selection with embedded color information to iteratively fine-tune the cascade network. 
However, despite the annotation on color information, this method requires a specific number of identities, which is hard to be known in real-life scenarios.

\section{Proposed Approach}

The pipeline of the proposed framework is shown in Fig.~\ref{fig:network}, which includes three parts:
1) viewpoint prediction, that identifies the viewpoint information through a viewpoint prediction network on input data, 2) recognition stage, that learns the discriminative feature for each sample using the repelled loss, and 3) progressive clustering, that uses the two-period algorithm to handle the problem of the similarity dilemma in clustering. 
The detailed optimization process is shown in Algorithm~\ref{algorithm}.

\subsection{Viewpoint Prediction.}
Due to the extreme viewpoint changes in vehicles, there are relatively small inter-class differences between different vehicles. We argue that global comparison in previous unsupervised clustering methods~\cite{lin2019bottom,ding2019dispersion,song2020unsupervised} tends to group the different vehicles with the same viewpoint into the same cluster.
Therefore, this global comparison scheme cannot guarantee the promising performance for target-only unsupervised vehicle Re-ID without any label supervision in network training.
To handle this problem, we propose to introduce a viewpoint prediction model to identify the vehicle's viewpoint information during the forthcoming clustering. 

In specific, we use a viewpoint prediction network to predict the viewpoint of each unlabeled vehicle image $x_i$ in training set $\left \{ X\mid  x_{1},x_{2},...,x_{N}\right  \}$.
%zheng-check and update
%
We train our viewpoint prediction model on VeRi-776~\cite{liu2016deep}, which contains all the visible viewpoints of the vehicle.
Following the viewpoints annotation in previous work~\cite{zheng2019attributes}, we divide vehicle images into five viewpoints, e.g., $v = \{front, front\_side, side, rear\_side, rear\}$.
Furthermore, we have additionally labeled 3000 samples in VeRi-Wild~\cite{lou2019large} data to fine-tune the model to improve the robustness of the viewpoint prediction. 
We use the commonly used cross-entropy loss $L_{\eta  }$ to optimize the viewpoint classifier $W\left (x_{i}\mid \theta \right)$,
\begin{equation}\label{loss_11}
L_{\eta  }=-\Sigma _{i}^{N}y_{v}log\left (W\left (x_{i}\mid \theta \right)\right)
\end{equation}
where $y_{v}$ is a one-hot vector of the ground truth of corresponding viewpoint labels.
%After passing this stage, all the data in the training set will be classified according to the viewpoint.
\label{VP}

\subsection{Recognition Stage.}
After the viewpoint prediction, we can obtain viewpoint-aware unlabeled  training set $X^{v} = \{x_{1}^{v},x_{2}^{v},...,x_{N}^{v}\}$, and the current training set can be regarded as the clusters divided according to the viewpoint. 
For example, VeRi-776~\cite{liu2016deep} falls into five different viewpoint clusters. 
%
%Therefore, we can 
%zheng- we can what?
%
For each image in $X^{v}$, we assign a unique 
index-label $y_{ind} = \{{1},{2},{3},...,{N}\}$ to indicate the category of each sample.
In order to learn the discriminative feature, one can achieve this objective by directly using triplet loss~\cite{cheng2016person,hermans2017defense} or cross-entropy loss via classification. 
However, the learning driven by these losses, which mainly calculate the similarity among each batch, will become inefficient and difficult to converge with the dataset's scale growth.  
Herein, we employ the more efficient repelled loss~\cite{xiao2017joint,zhong2019invariance,lin2019bottom}, which calculates the feature similarity between the current sample and all the training samples at once.

It is equipped with a key-value structure to store the features of all training samples, and the index-label $y_{ind}$ is stored in the key memory. 
The $y_{ind}$ will not change during the entire training process. 
We calculate the feature similarity between the $i$-th image in the $v$-th viewpoint $f_{i}^{v}$ and all the samples,
%
%, and it should be noted that the index-label $y_{id}$ not the pseudo label $y_{p}$ of each sample. But since the purpose of this stage is to make the model distinguish each sample as much as possible, so the $y_{id}$ is equal to the $y_{p}$ in value. Because there is a one-to-one correspondence between key and value, the index-label can uniquely determine the sample, so we can assign $y_{p}$ based on the $y_{id}$. We calculate the similarity between $f$ and all samples stored in the feature memory,
\begin{equation}\label{loss_1}
p(y_{p} |x_{i}^{v}) =\frac{\exp(\left( M[ i]^{T} f_{i}^{v} /\beta \right)}{\sum ^{N}_{j=1}\exp(\left( M[ j]^{T} f_{i}^{v} /\beta \right)}
\end{equation}
where $M[i]$ denotes the $i$-th slot of the value memory $M$.
$\beta$ is a hyper-parameter to control the softness of the probability distribution over classes, which is set to 0.1 followed by~\cite{lin2019bottom}. 
$N$ indicates the number of clusters.
$y_{p}$ is the pseudo label, and we initialize $y_{p}=y_{ind}$. 
We maximize the distance between samples by assigning each sample to its own slot,
\begin{equation}\label{loss_2}
L_{\alpha}=-\log(p(y_{p} |x_{i}^{v}))
\end{equation}

During the back propagation, the feature memory is updated by the formula $M[y_{i}]\leftarrow\tfrac{1}{2}( M[ y_{i}]+f_{i}^{v})$. 
At the recognition stage, $M[ y_{i}]$ stores the features of each training sample.
At the subsequent progressive clustering stage, the pseudo label $y_{p}$ of each sample will be redistributed according to the clustering results, while each slot stores the features of each cluster.
\label{recognition}

\subsection{Progressive Clustering.}
Without any identification information, we propose a progressive clustering algorithm for unsupervised vehicle Re-ID.
It mainly contains three aspects, two-period algorithm to avoid the similarity dilemma caused by the extreme viewpoint changes of vehicles, the $k$-reciprocal encoding to re-metric the distance for more robust clustering, and clustering with noisy sample selection to deal with outliers that are difficult to be clustered in real scenes.

{\flushleft \bf The first Period}.
Through the recognition stage, the model learned more recognizable identity features of each image.
%zheng-which stage? why pay attention to......?
%
The features obtained from the training set $F^{*} = \{F_{1}, F_{2}, F_{v}, ..., F_{V}\}$,
\begin{equation}
  F_{v} = \{f_{1}^{v},f_{2}^{v},...,f_{N_{v}}^{v}\}
\end{equation}
%zheng-simplify to one equation
%
where $F_{v}$ and $N_{v}$ represent the feature set and the number of samples in the $v$-th viewpoint.
We compare the similarity of all features $F_{v}$ belonging to the same viewpoint cluster to obtain the distance matrix $D( F_{m} ,F_{n}) ,\ m=n$.
$D$ represents the scoring matrix of Euclidean distance $d_{ij} =\| f_{i} -f_{j} \| ^{2}$. 
There is no doubt that the same vehicle with the same viewpoint has the highest similarity and thus tends to be clustered together (assigned to the same pseudo label) with the highest priority.
For each different distance matrix in the same viewpoint, we obtain pseudo labels by the prevalent cluster algorithm DBSCAN~\cite{ester1996density}, which can effectively deal with noise points and achieve spatial clusters of arbitrary shapes without information of the number of clusters compared to the conventional k-means~\cite{kanungo2002efficient} clustering.

{\flushleft \bf The Second Period}.
In the second period, we compare the distance between different viewpoint clusters.
%zheng-constrain what? 
%
We take the shortest distance between features in two clusters as a measure of the distance between clusters.
Considering that we have no idea whether the current sample has positive samples (with the same identity) in other viewpoints,
we comprehensively compare the distance between all different viewpoint clusters,
\begin{equation}
  D^{*}_{mn}=\left \{ D(F_{1},F_{2}), ..., D(F_{m},F_{n}) \right \},\ m\neq n.   
\end{equation}

We argue that the higher similarity, the more likely the same identity. 
Thus adopt a progressive strategy to merge the clusters between different viewpoints gradually.  
Therefore, We first calculate a rank list $R$,  
%zheng-why?
%
\begin{equation}\label{R1} 
R=argsort(D_{mn}^{*}),m\neq n
\end{equation}
where $R$ finds the most similar clusters among all different viewpoints.
%
%zheng-hard to follow
We set a strict distance threshold $\tau $,
and merge clusters from different viewpoints only when the distance of the candidates in $R^{*}$ is less than $\tau $, i.e,
\begin{equation}\label{R2} 
R^{*}=R[1:C(d=\tau )]
\end{equation}
\vspace{-0.7cm}

\noindent
where  $C=\{c_{i},c_{j}\}$ is the last sample pair in different clusters with distance less than $\tau$.
Intuitively, due to the style diversity of different datasets, we expect the setting of $\tau $ to be irrelevant to datasets. 
In our method, after the recognition stage, we ascending sort the calculated $D^{*}_{mn}$, and set the distance value between the $ti$-th lowest sample pair as the threshold $\tau $. 
The distance threshold is only calculated after the recognition stage and then fixed in the whole training process. 
We alternately execute the above two periods during each iteration. The model learns the features of vehicles from the same viewpoint while continuously mining the features of vehicles with the same identity from different viewpoints.
\begin{algorithm}[t]
\caption{The viewpoint-aware progressive clustering method (VAPC)}
\label{algorithm}
\begin{algorithmic}[1]
\REQUIRE
Unlabeled training set $X = \{x_{1},x_{2},x_{3}...,x_{N}\}$;
Recognition stage epoch $E_{r}$;
Set the distance of the most similar $ti$-th sample pair as the distance threshold;
CNN model $M$; 
%zheng-indicate what is this CNN model.
%
index-label $y_{ind}=1,2,3,...,N$. \\
%\STATE pseudo label $y_{p}$=$y_{id}$. \\
\STATE Viewpoint prediction: $X$ $\rightarrow$ $X^{v}$, $V = 5$;
%zheng-???
%
\STATE Recognition stage: \\
\FOR {$ i < E_{r}$} 
\STATE Train CNN model $M$ with $X$ and $y_{ind}$ according to Eq.~\eqref{loss_2}. \\
\ENDFOR
\STATE Calculate threshold $\tau $; \\
\ENSURE
Best CNN model $M$\\
\STATE Progressive clustering stage: \\
\STATE First period:\\
\FOR {$ i < V$} 
\STATE Calculate distance matrix: $D (F_{i} ,F_{i} )$.\\
%zheng-???check
%
\STATE Re-metric distance by Eq.~\eqref{diss_2} to obtain $D_{J}(F_{i} ,F_{i})$
\STATE Use DBSCAN to obtain clustering results. 
\ENDFOR
\STATE Mine noise samples according to Eq.~\eqref{noise_sel}. \\
\STATE Second period: \\
\STATE Compare feature sets at the different viewpoint to obtain distance matrix $D( F_{m} ,F_{n} ), m\neq n$. 
\STATE Select the clusters need merged from different viewpoints through Eq.~\eqref{R1} and Eq.~\eqref{R2}. \\
\STATE Retrain CNN model $M$ with $X$ and $y_{p}$ according to Eq.~\eqref{loss_2}. \\
\STATE Evaluate on the test set $\rightarrow$ performance $P$ \\
\IF{$P > P^{ \ast }$} 
\STATE $P^{ \ast } = P$ 
\STATE Save the best model $M$
\ENDIF
\end{algorithmic}
\end{algorithm}
{\flushleft \bf Distance metric by $k$-reciprocal encoding}.
Clearly, more positive samples in the same-viewpoint cluster in the first period, higher clustering quality at different viewpoints in the second period, which in turn will benefit the performance in the next iteration.  
%the clustering quality of the same viewpoint in the first period is crucial in our method. 
%
%
Note that the clustering method significantly relies on the distance metric, we propose introducing the widely used $k$-reciprocal encoding~\cite{zhong2017re,Fu_2019_ICCV,song2020unsupervised} as the distance metric for feature comparison. 
For the sample $x_{i}^{v}$ in $X^{v}$, we record its $k$ nearest neighbors with index-labels $K_{k}(x_{i}^{v})$, for all indexes $ind\in K_{k}(x_{i}^{v})$, if $ \left | K_{k}(_{i}^{v}) \cap K_{\frac{k}{2}}(x_{ind}^{v}) \right  |\geqslant \frac{2}{3} \left | K_{ \frac{k}{2}}(x_{ind}^{v}) \right | $, $x_{i}^{v}$'s mutual $k$ nearest neighbors set $ S_{i} \leftarrow \left | K_{k}(x_{i}^{v}) \cup   K_{\frac{k}{2}}(x_{ind}^{v}) \right  | $. 
%zheng-update
%
In this case, all reliable samples similar to $x_{i}^{v}$ are recorded in $S_{i}$. 
%zheng-hard to follow
%
Then distance $d_{ij}$ of the sample pair in the same viewpoint distance matrix, $D( F_{m} ,F_{n}) ,\ m=n$ reassigns weight by,
%zheng-update what?
%
\begin{equation}\label{diss_1}
\tilde{d}_{ij}=\left\{
\begin{aligned}
&e^{-d_{ij}}\quad if \ j\in S_{i},\\
&0\quad \quad \ else
\end{aligned}
\right.
\end{equation}

For each image pairs $(x_{i}^{v},x_{j}^{v})$ at the same viewpoint, we get a new distance matrix $D_{J}( F_{m} ,F_{n}) ,\ m=n$ for clustering, it can be calculated by,
\begin{equation}\label{diss_2}
d_{J}(x_{i}^{v},x_{j}^{v})=1-\frac{\sum_{l=1}^{N_{v}}min(\tilde{d}_{il},\tilde{d}_{jl})) }{\sum_{l=1}^{N_{v}}max(\tilde{d}_{il},\tilde{d}_{jl}))  }
\end{equation}
where $N_{v}$ is the total number of samples in viewpoint $v$. 

\begin{figure}[t]
\centering
\includegraphics[height=5.4cm,width=8.9cm]{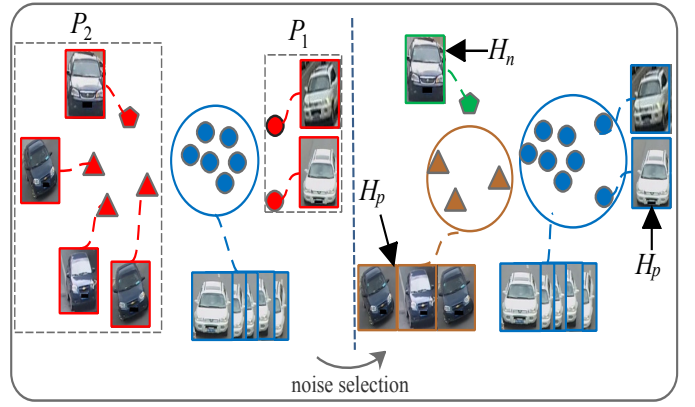}
%\caption{fig1}
\caption{ Illustration of noise selection. Samples in the same color belong to the same cluster except the red color is for noisy samples (pseudo label is -1). $P_{1}$, $P_{2}$ represent two different noise situations. After noise selection, we reconstruct the cluster for the noise samples by comparing each noise and other clusters. $H_{p}$ and $H_{n}$ represent hard positive samples and hard negative samples, respectively.}
\label{yczs}
\vspace{-0.3cm}
\end{figure}
{\flushleft \bf Clustering with noisy sample selection}.
Our viewpoint-aware clustering strategy avoids comparing different viewpoints of vehicles during the first period of clustering, which alleviates the intra-class gap and reduces the difficulty of clustering to a great extent.
However, due to the complexity of the real scene, some hard samples are still difficult to cluster and then regarded as noises.
The reason is, although DBSCAN~\cite{ester1996density}  can generate clusters for data of any spatial shape,
%zheng-first admit the advantage
%
it uses two parameters $eps$ and $minPts$ to define the density conditions that need to be meet when forming clusters in the training set, which tends to cluster the samples with small intra-class gaps and treat the samples with larger intra-class gaps as noises, 
as shown in Fig.~\ref{yczs}. 
We observe that these noises usually derive from two situations which are shown as $P_{1}$ and $P_{2}$ in Fig.~\ref{yczs}.
In $P_{1}$, due to occlusion, misalignment of the bounding box, or deviation of the viewpoint prediction, samples with same identity but far from the already formed clusters (the blue cluster as shown in Fig.~\ref{yczs}), will be regarded as noise.
In $P_{2}$, some samples deriving from the same identity fail to form into the same cluster since they can not meet the density condition due to large intra-class differences. 
%
%
%Based on the analysis of the formation of the above two kinds of noisy samples, we think that each noisy sample has two kinds of similarity relationships. 
%基于以上两种噪声形成的分析，我们认为每一个噪声样本都有两种相似性关系

For $P_{1}$, we expect noises samples can be classified into clusters with the same identity. For $P_{2}$, we expect that noise samples with the same identity can be clustered together to form a new cluster.
%对于第一种情况我们依据相似性将噪声点和集群进行合并，第二种情况我们将具有高相似性的噪声点进行合并
Specifically, we use set $S_{n}$ to collect all the noise samples. For each member $s_{i}$ in $S_{n}$, we look for the most similar samples with the same viewpoint $p_ {i} $, and constitute a set of similar sample pairs, $\{\{s_{1},p_{1}\},\{s_{2},p_{2}\},...,\{s_{n},p_{n}\}\}$, which is descending sorted according to their pairwise similarities.
Then we judge which situation the noise belongs to based on $p_{i}$. 
If $p_{i}$ belongs to a cluster that has already been clustered, it corresponds to the first situation $P_{1}$.
Otherwise, $p_{i}$ is a noise sample, it belongs to the second situation $P_{2}$. 
%根据{xi,pi}图像对的相似性从高到低，如果pi是属于已经被聚类的集群那么属于第一种情况，如果p是噪声那么是第二种情况
However, directly merging $\{s_{i},p_{i}\}$ is not reliable caused by some hard negative samples, we take a more reliable approach as follows.
%并不是直接将他们合并，受kR启发，我们采取了一种更可靠的做法，如下
Inspired by $k$-reciprocal encoding, if $\{s_{i},p_{i}\}$ belong to the same identity, their neighbor image sets should be similar, which also means that they should be located in each other's $\tilde{k}$-nearest neighbors.
%如果\{x_{i},p\是属于同一身份的，那么他们的邻居样本集合应该是一样的。也就是说如果他们属于同一身份，那么他们应该是对方的k近邻的邻居集合中
Therefore, we calculate the $\tilde{k}$-nearest neighbor image sets of the same viewpoint as $p_{i}$.
%因此我们计算p的k近邻集合$top\tilde{k}[p_{i}]$
If $s_{i}$ appears in $top\_\tilde{k}[p_{i}]$, $s_{i}$ is regarded as a reliable hard positive sample and will be merged with $p_{i}$.
Otherwise, the noisy sample $s_{i}$ will be treated as a hard negative sample and divided into a new cluster to further learn its discriminative feature.
%如果xi在$top\tilde{k}[p]$中则xi是困难正本，将他们合并，如果不在，噪声样本xi将被视为困难负样本，分为一个单独的集群以进一步学习其判别性特征。
Formally, we construct:
\begin{equation}\label{noise_sel} 
\left\{
             \begin{array}{lr}
             H_{p_{1}}=\left \{\left ( s_{i},p_{i} \right ) \mid p_{i} \notin S_{n}, top\_1[s_{i}]=p_{i}, s_{i} \in top\_\tilde{k}[p_{i}]  \right \} &  \\
			 H_{p_{2}}=\left \{\left ( s_{i},p_{i} \right ) \mid p_{i} \in S_{n}, top\_1[s_{i}]=p_{i},  s_{i} \in top\_\tilde{k}[p_{i}]  \right \} &  \\
             H_{n}=\left \{ \left ( s_{i} \right ) \mid top\_1[s_{i}]=p_{i}, s_{i} \notin top\_\tilde{k}[p_{i}] \right \} &  \\
 
             \end{array}
\right.
\end{equation}

We merge $s_{i}$ in $H_{p1}$ with the corresponding $p_{i}$, in $H_{p_{2}}$, we form a new cluster $C_{ni}$ for each pair $(s_{i}, p_{i})$, and treat each hard negative sample in $H_{n}$ as a single cluster. Note that when $C_{ni}$ be created, for $c_{i}$ in $C_{ni}$, we will dynamically look for $top\_\tilde{k}[c_{i}]$ as candidates and determine whether to merge the candidate with $C_{ni}$ according to the condition of $H_{p_{2}}$, and not merge with other clusters after merging with $C_{ni}$.
\label{pc1}

\section{Experiments}
We evaluate our proposed method VAPC on two benchmark datasets VeRi-776~\cite{liu2016deep} and VeRi-Wild~\cite{lou2019large}, which contain $5$ and $4$ view-points respectively.
We compare our method with the prevalent domain adaption based unsupervised, and target-only methods without domain adaption for evaluation.

\subsection{Datasets and Evaluation Protocol.}
%We evaluate our method on two multi-view benchmark datasets, VeRi-776~\cite{liu2016deep} and VeRi-WILD~\cite{lou2019large} which contain $5$ and $4$ view-points respectively.
%zheng-update
%
%Our goal is to bring positive samples together by measuring the similarity between vehicle viewpoints. It should be noted that the two viewpoint, which have no intersection at all, have almost no similarity without label guidance, e.g., side and front or front and back. In this way our experiments are mainly conducted on multi-view dataset.

\noindent
\textbf{VeRi-776}~\cite{liu2016deep} is a comprehensive vehicle re-identification dataset providing rich attributes information such as color, type and temporal path. 
It contains 776 different vehicles captured in 20 cameras, yielding more than 49,357 images and 9,000 tracks.
The training and testing sets contain 37,728 images of 576 vehicles and 11579 images of 200 vehicles.
Both training and testing sets contain $5$ common visible viewpoint situations, including $front, front\_side, side, rear\_side, rear$.
%zheng-indicate we obtain this annotation from xianming's paper.
%
Following the protocol in~\cite{liu2016deep}, we only return the matchings from the different cameras for the query vehicles as the results.
We use the mean average precision (mAP) and cumulative matching characteristic (CMC) at Rank-1, Rank-5 and Rank-20 as the measurement metrics.

\noindent
\textbf{VeRi-Wild}~\cite{lou2019large} is a large-scale vehicle Re-ID dataset, containing more than 400 thousand images of 40 thousand vehicle IDs captured by 174 cameras in the surveillance system.
%zheng-VERI-Wild? VeRi-WILD,keep consistent all through the paper.
%
It contains complex backgrounds, various viewpoints and illumination variations in real-world scenes. 
%zheng-indicate how to obtain the viewpoint labels
%
The training set contains 277,797 images of 30,671 vehicles.  After the viewpoint prediction of the training set, VeRi-Wild contains 4 viewpoints, $front, front\_side, rear\_side, rear$, respectively containing 110204, 52716, 64968, 49909 images. Due to hardware limitations, we use all the training data in the recognition stage, and each viewpoint in the clustering stage takes 10,000 images, respectively.
While the testing set consists of three subsets, test-3000, test-5000, and test-10000 with different testing sizes. 
Following the protocol in~\cite{lou2019large}, the match rate protocol on VeRi-Wild is that all the references of the given query are in the gallery.
%zheng-can not understand this sentence
%
We use mAP, Rank-1 and Rank-5 as the evaluation metrics.
\begin{table*}[htbp]
\caption{Comparison with the state-of-the-art of target-only Re-ID and domain adaptive Re-ID methods on VeRi-776 and VeRi-Wild. “src” denotes the source domain/dataset, where “N/A” and "VehicleID" indicate the target-only  and domain adaptive methods on VehicleID dataset respectively. "VAPC\_TO", "VAPC\_DT" and "VAPC\_DA" indicate our VAPC in target-only, direct transfer and domain adaption respectively.}
\resizebox{\textwidth}{23mm}{
\setlength{\tabcolsep}{2.7mm}{
\begin{tabular}{c|c|cccc|ccc|ccc|ccc}
\hline
\multirow{3}{*}{method} & \multicolumn{5}{c|}{VeRi-776}                                                                                  & \multicolumn{9}{c}{VeRi-Wild}                                                                                         \\ \cline{2-15} 
                        & \multirow{2}{*}{src} & \multirow{2}{*}{R1} & \multirow{2}{*}{R5} & \multirow{2}{*}{R20} & \multirow{2}{*}{mAP} & \multicolumn{3}{c|}{test-3000} & \multicolumn{3}{c|}{test-5000} & \multicolumn{3}{c}{test-10000} \\ \cline{7-15} 
                     &                     &                     &                      &                      &                      & R1       & R5       & mAP      & R1       & R5       & mAP      & R1       & R5       & mAP      \\ \hline
OIM~\cite{xiao2017joint}                 & N/A                  & 45.1                & 62.2                & 78.1                & 12.2                  &     48.7     &   66.6       &     14.4     &     45.0      &     60.9     &    12.6 &     38.8     &    54.4      &   10.0       \\
Bottom~\cite{lin2019bottom}                 & N/A                  & 63.7                & 73.4                & 83.4                 & 23.5                  &     70.5     &   86.0       &     30.7     &     64.2     &     82.2     &    27.1      &     55.2     &    75.1      &   21.6       \\
AE~\cite{journals/tomccap/DingFXY20}                     & N/A                  & 73.4                &  \bf 82.5                & \bf 89.7                 & 26.2                  &     68.5     &     87.0     &  29.9        &     61.8     &     81.5     &   26.2       &     53.1     &    73.7      &    20.9      \\ \hline
{\bf VAPC\_TO (ours)}              & N/A                  &  \bf 76.2                &   81.2                 &    85.3                  &  \bf 30.4                  &  \bf 72.1        &   \bf 87.7      &    \bf 33.0      &     \bf 64.3     & \bf 83.0         &  \bf 28.1        &  \bf 55.9       &    \bf 75.9      &    \bf  22.6     \\ \hline
SPGAN~\cite{deng2018image}                   & VehicleID            & 57.4                & 70.0                & -                    & 16.4                 &     59.1     &     76.2     &     24.1     &     55.0     &   74.5       &    21.6      &     47.4     &     66.1     &    17.5      \\
ECN~\cite{zhong2019invariance}                    & VehicleID            & 60.8                & 70.9                & 85.4                 & 27.7                  &   73.4       &     88.8     &     34.7     &   68.6       &     84.6     &    30.6      &   61.0       &     78.2     &    24.7      \\
UDAP~\cite{song2020unsupervised}                    & VehicleID            & 76.9                & \bf 85.8                & -                    & 35.8                  &     68.4    &     85.3     &    30.0      &    62.5      &     81.8     &    26.2      &     53.7     &       73.9 &     20.8 \\ \hline
{\bf VAPC\_DT (ours)}   & VehicleID            & 69.1                & 79.0                & 88.2                 & 35.5                 &      74.0    &     88.6     &    37.7      &   68.1       &    84.8      &    33.1      &    60.2      &    78.7      &    26.3      \\
{\bf VAPC\_DA (ours)}              & VehicleID            &        \bf 77.4                &   84.6                &  \bf 91.6                 & \bf 40.3                 &     \bf 75.3     &   \bf 89.0      &   \bf 39.7       &  \bf  69.0      &   \bf 85.5      &  \bf 34.5      &  \bf 61.0      &  \bf  79.7      &  \bf 27.4    \\ \hline
\end{tabular}}}
\label{sota}
\vspace{-0.3cm}
\end{table*}
\begin{figure*}[t]
\centering
\subfigure[OIM]{
\begin{minipage}[t]{4.35cm}
\centering
\includegraphics[height=4cm,width=4cm]{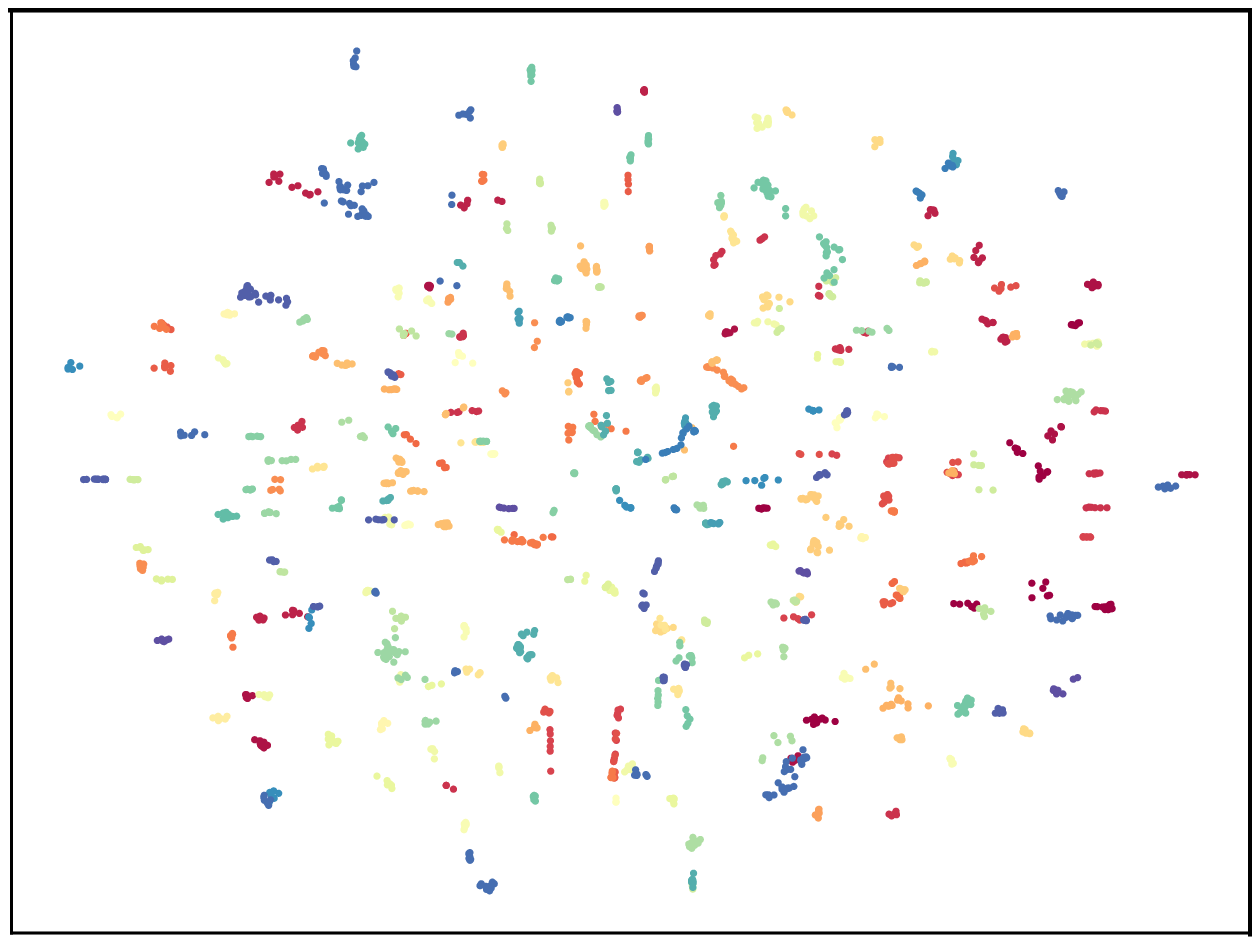}
%\caption{fig1}
\end{minipage}%
}%
\subfigure[Bottom]{
\begin{minipage}[t]{4.35cm}
\centering
\includegraphics[height=4cm,width=4cm]{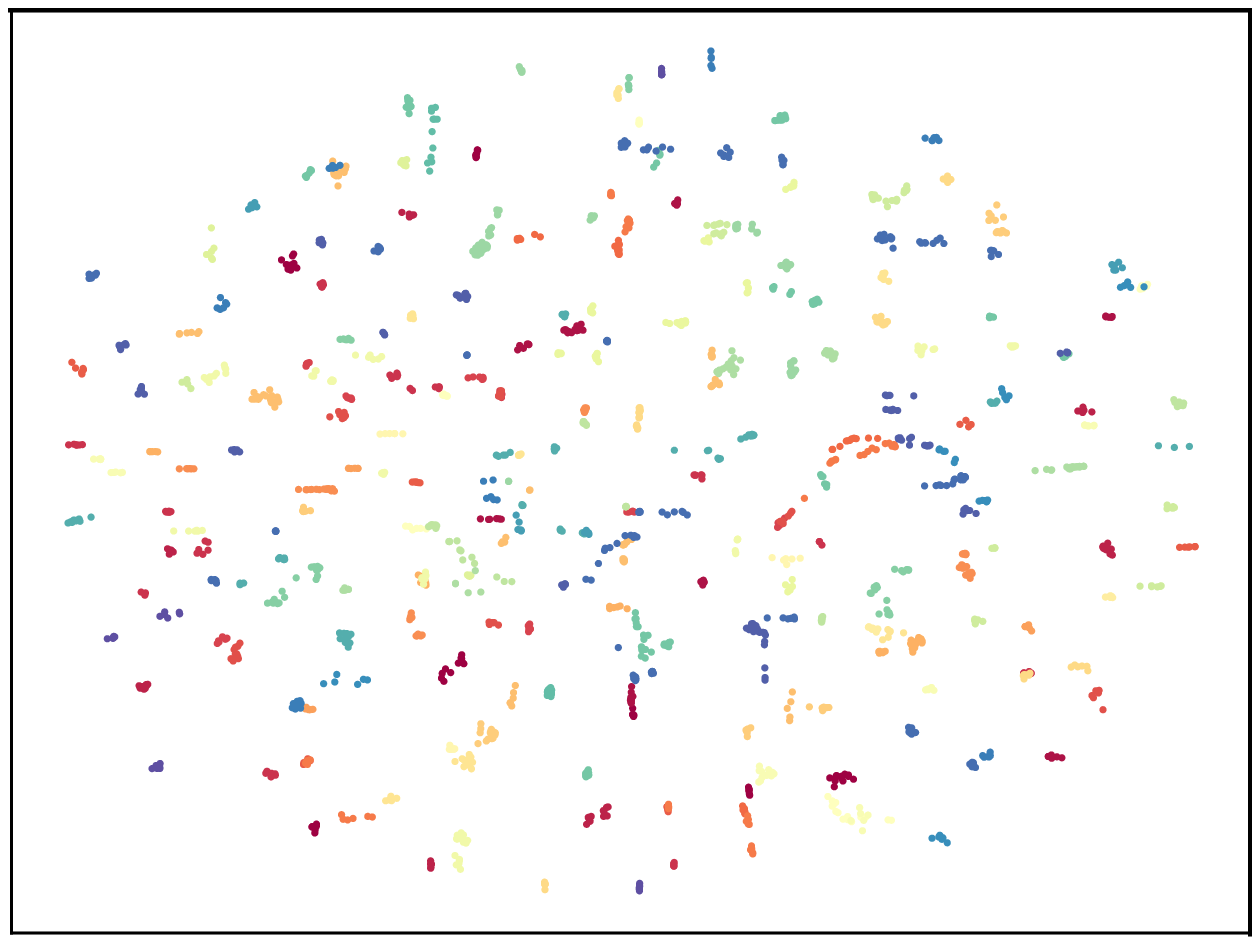}
%\caption{fig2}
\end{minipage}%
}%
\centering
\subfigure[AE]{
\begin{minipage}[t]{4.35cm}
\centering
\includegraphics[height=4cm,width=4cm]{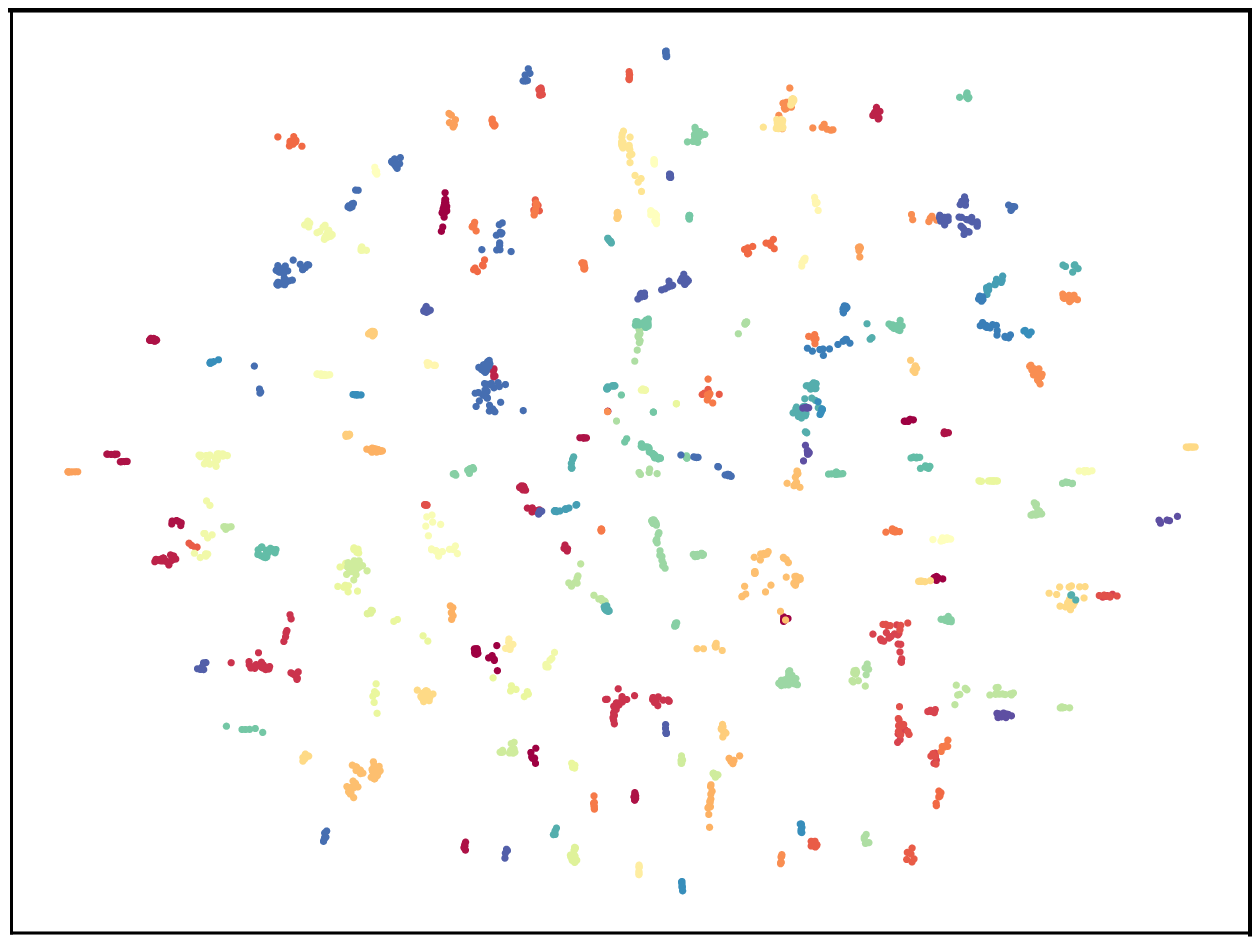}
%\caption{fig1}
\end{minipage}%
}%
\subfigure[VAPC\_TO (ours)]{
\begin{minipage}[t]{4.35cm}
\centering
\includegraphics[height=4cm,width=4cm]{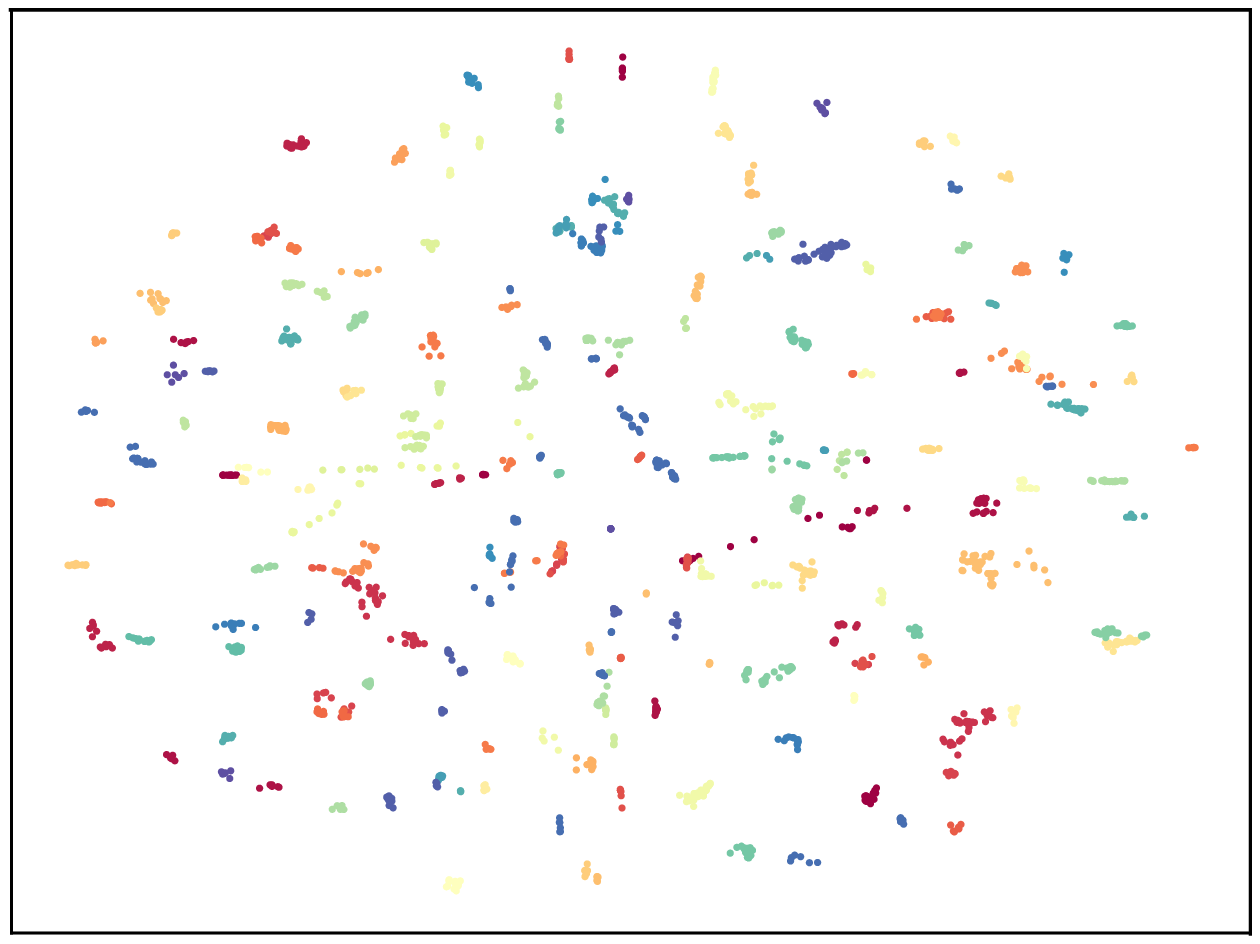}
%\caption{fig2}
\end{minipage}%
}%
\caption{ Visualization for features extracted by target-only  method, OIM~\cite{xiao2017joint}, Bottom~\cite{lin2019bottom}, AE~\cite{journals/tomccap/DingFXY20} and our method. 37 identities with 2000 images in the gallery of VeRi-776 are used. Each point represents an image, and each color represents a vehicle identity.}
\label{vis}
\vspace{-0.3cm}
\end{figure*}

\noindent
\textbf{VehicleID}~\cite{liu2016deep2} is another large-scale vehicle Re-ID dataset, it includes 110,178 real scene images of 13134 types of vehicles 
as a training set. 111,585 images of 13,113 vehicles were used as a test set. In this article, to compare the results of other existing unsupervised domain adaptation methods, we also use the VehicleID dataset as the source domain for supervised training. 

\noindent
\subsection{Implementation Details.}
We use ResNet50~\cite{he2016deep} as the backbone by eliminating the last classification layer.
All experiments are implemented on two NVIDIA TITAN Xp GPUs. 
We initialize our model with pre-trained weights on ImageNet~\cite{deng2009imagenet}. 
For the viewpoint prediction network, we set the batch size as 32 and the learning rate as 0.001, with a maximum 20 epochs. 
If not specified, we use stochastic gradient descent with a momentum of 0.9 and the dropout rate as 0.5 to optimize the model. 
For the Re-ID feature extraction network, we resize the input images of VeRi-776~\cite{liu2016deep} and VeRi-Wild~\cite{lou2019large} as (384,384).
The batch size is set to 16. The learning rates at the recognition stage is set to 0.1 and divided by 10 after every 15 epochs, and set to 0.001 in the clustering stage. We only use a random horizontal flip as a data augmentation strategy. Following the protocol in~\cite{zhong2017re} we set $k$ to 20.
%
%for the DBSCAN algorithm, we empirically set the parameter $minPts$ to 4, and the $eps$=$average(D(R_{v}[1:\gamma]))$. $D(R_{v}[1:\gamma])$ means the distance between the most similar $\gamma$ sample pairs in the feature set of the viewpoint $v$. 
%zheng-why introduce DBSCAN? what are these parameters meaning? make no sense.
%Following the protocol in~\cite{song2020unsupervised}, we set $\gamma=1.6\times 10^{-3}N$, where $N$ represents the number of samples in the current feature sets. 
%zheng-what is \gamma?
%Since VeRi-Wild has a smaller inter-class distance compared to VeRi-776, we set a smaller $\gamma=5\times 10^{-4}N$. 
%
%We empirically set the $ti$ =1200 and the $\tilde{k}$ = 1. 
%zheng-what the meaning of $ti$ and $\tilde{k}$ ?
%

%The training of the source domain converges after 20 epochs.
%

\subsection{Comparison with State-of-the-art Methods.}
We compare our method with the state-of-the-art unsupervised Re-ID methods on VeRi-776~\cite{liu2016deep} and VeRi-Wild~\cite{lou2019large} in both target-only and domain adaption scenarios, as shown in TABLE~\ref{sota}. 

\noindent
{\bf Compared with the target-only method.}
We first compare our method with three state-of-the-art target-only unsupervised methods OIM~\cite{xiao2017joint}, Bottom~\cite{lin2019bottom} and AE~\cite{journals/tomccap/DingFXY20}.
Generally speaking, our method (VAPC\_TO) outperforms the three state-of-the-art target-only methods by a large margin by exploring the intra-class relationship.
OIM~\cite{xiao2017joint} devotes to extracting discriminative features efficiently, which ignores the intra-class relationship, thus results in stumbling performance.
Bottom~\cite{lin2019bottom} designs a bottom-up clustering strategy by merging the fixed clusters during each step.
However, each clustering may produce the wrong classification, and more clustering steps, more clustering errors.
Especially on the VeRi-776~\cite{liu2016deep}, almost all visible viewpoints are included, which brings greater clustering challenges. Each clustering step only focuses on the same viewpoint and can not bring more samples of different viewpoints together. Our method effectively alleviates this problem and brings greater improvement.
%zheng-update the reason
%
AE~\cite{journals/tomccap/DingFXY20} clusters the samples via a similarity threshold and constrains the cluster size by embedded a balance term into the loss. 
However, due to the similarity dilemma of vehicles, where the same viewpoints of different identities may have higher similarities, it is difficult to set an optimal similarity threshold for clustering. 
In addition, more and more samples meeting the similarity threshold are treated as the same identity during the training,
especially on larger scaled dataset VeRi-Wild~\cite{lou2019large}, it will cause more severe data imbalance in each cluster and damage the feature representation.
Therefore the performance of AE~\cite{journals/tomccap/DingFXY20} on VeRi-Wild~\cite{lou2019large} declines comparing with Bottom~\cite{lin2019bottom}.
%zheng-update
%

%
%Comparing with VeRi-Wild, our method achieves much higher improvement on VeRi-776 dataset, the main reason is that compared to VeRi-776, VeRi-Wild has fewer samples with the same identity in different viewpoints.
%zheng-update

We further use t-SNE~\cite{dermaaten2008visualizing} to visualize the feature space distribution of our method compared to the three state-of-the-art target-only methods, as shown in Fig.~\ref{vis}.
Compared with ours, the distribution between the points is sparser in  OIM~\cite{xiao2017joint} and Bottom~\cite{lin2019bottom}, while more points of different colors gathering in AE~\cite{journals/tomccap/DingFXY20}.
our method presents a better feature distribution, which demonstrates that VAPC\_TO can successfully cluster more images of vehicles with the same identity and effectively improve the feature representation for unsupervised vehicle Re-ID.

\noindent
{\bf Compared with unsupervised domain adaptation.}
To evidence the effectiveness of our method on unsupervised vehicle Re-ID, we further evaluate our method in the domain adaption fashion.
Following the protocol in~\cite{song2020unsupervised}, we use VehicleID~\cite{liu2016deep2} as the source domain and employ repelled loss~\cite{lin2019bottom} for supervised training, replacing the recognition stage in~\ref{recognition}. 
We compare our method in the domain adaption fashion (VAPC\_DA) with three state-of-the-art unsupervised domain adaptation methods, including SPGAN~\cite{deng2018image}, ECN~\cite{zhong2019invariance} and UDAP~\cite{song2020unsupervised}, as shown in the lower half part in TABLE~\ref{sota}.

SPGAN~\cite{deng2018image} considers the style change among different datasets and trains a style conversion model to bridge the style discrepancy between the source domain data and the target domain. 
However, due to the huge gap between the vehicle datasets in the real scene, e.g., the diverse viewpoints, resolution and illumination, it is challenging to obtain the desired translated image, which is crucial in SPGAN~\cite{deng2018image}, and thus results in poor performance for vehicle Re-ID. 
ECN~\cite{zhong2019invariance} joins the source domain for model constraints while using the $k$-nearest neighbor algorithm to mine the same identity in the target domain. 
The setting of the $k$ value not only has a greater impact on the experimental results, but the most similar top $k$ samples are always at the same viewpoint.
%zheng-hard to understand
%zheng-explain why ECN doesn't work individually first,
%
UDAP~\cite{song2020unsupervised} uses source domain data to initialize the model and theoretically analyzes the rules that the model needs to follow when adapting to the target domain from the source domain.
It achieves satisfactory results on vehicle Re-ID due to the strengthening of the constraints on the target domain training. The target domain feature extractor has stronger learnability while obtaining the source domain knowledge.
However, it relies on global comparison, which may cause more clustering errors, especially on VeRi-Wild~\cite{lou2019large} dataset presents a much smaller inter-class differences than VeRi-776~\cite{liu2016deep}.
\begin{table}[t]
\caption{Results evaluated on the VeRi-776 and test-3000 set of VeRi-Wild. kR means distance metric by $k$-reciprocal encoding, NS means noise selection, FS represents our first period and second period clustering strategy.}
\resizebox{88mm}{15mm}{
\begin{tabular}{l|c|c|c|c|c|c}
\hline
\multirow{3}{*}{method} & \multicolumn{3}{c|}{VeRi-776}                                        & \multicolumn{3}{c}{VeRi-Wild}                                       \\ \cline{2-7} 
                        & \multirow{2}{*}{R1} & \multirow{2}{*}{R5} & \multirow{2}{*}{mAP} & \multirow{2}{*}{R1} & \multirow{2}{*}{R5} & \multirow{2}{*}{mAP} \\
                        &                       &                       &                      &                       &                       &                      \\ \hline
(a) Ours                  & 76.2                 & 81.2                  & 30.4                 &     72.1            &          87.7           &     33.0               \\ 
(b) w/o tP                & 68.7                 &  73.2                 & 25.0                &         68.5 &           85.0       &               30.3 \\
(c) w/o kR              & 71.0                  & 78.9                  & 24.1                 &       69.2           &       86.0            &        29.7          \\
(d) w/o NS       &       71.3                 &     78.8                 &      27.8                 &        70.1               &    87.0                   &    32.5                  \\
(e) w/o tP + kR + NS  & 61.4                  & 72.5                  & 18.2           &   48.7              &           66.6        & 14.4                 \\
\hline
\end{tabular}}
\label{abl}
\vspace{-0.3cm}
\end{table}
%Both ECN~\cite{zhong2019invariance} and UDAP~\cite{song2020unsupervised} work overshadowed comparing to our method.
%
\begin{figure}[t]
\centering
\includegraphics[height=16cm,width=8.5cm]{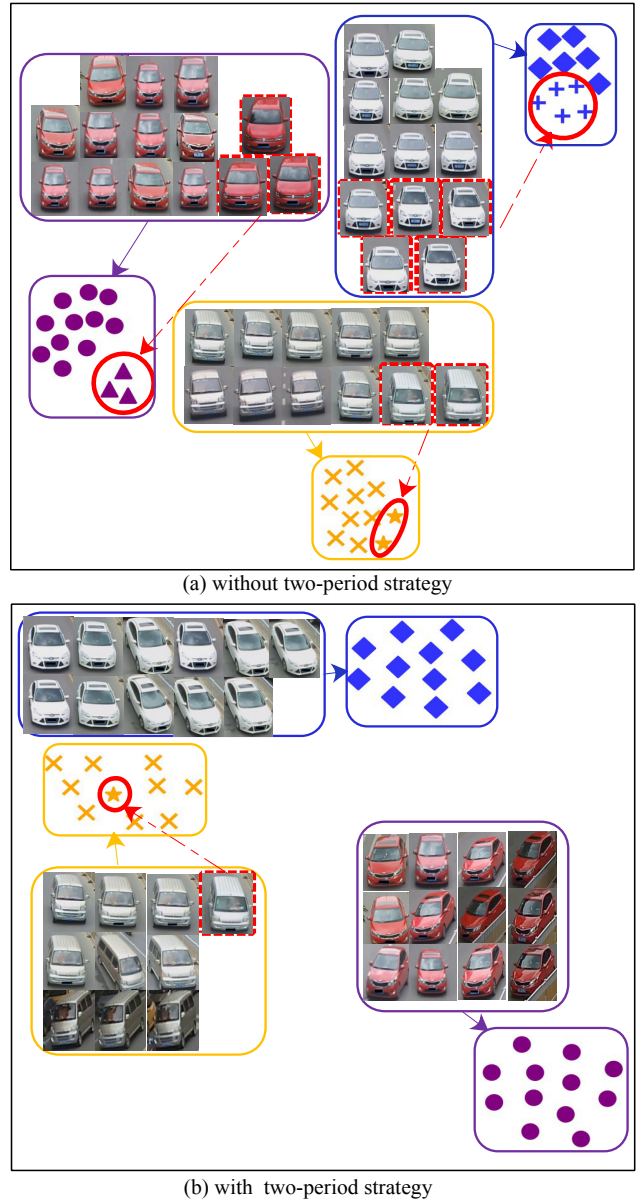}	
\caption{Illustrations with and without two-period strategy on VeRi-776. The same color represents the same cluster, and different shapes represent different identities. The red circle marks the false clustered samples.}
\label{fs}
\vspace{-0.4cm}
\end{figure}

In addition, we evaluate our method in the "Direct Transfer" fashion by training on the source domain and directly testing on the target domain indicated as (VAPC\_DT) in TABLE~\ref{sota}.
First of all, by leveraging the information in the training data, VAPC\_DT generally outperforms VAPC\_TO, which verifies the knowledge of the source domain during the training improves the vehicle retrieval ability of the model.
The only exception is the rank-1 score on VeRi-776~\cite{liu2016deep}.
The main reason is the huge gap between VehicleID~\cite{liu2016deep2} and VeRi-776~\cite{liu2016deep} datasets, e.g., VeRi-776 has lower resolution and more viewpoints, which results in poor generalization performance.
Even though VAPC\_DT significantly boosts the mAP score on both VeRi-776~\cite{liu2016deep} comparing to the target-only fashion (VAPC\_TO).
Second, VAPC\_DT is even significantly superior to the domain adaption methods SPGAN~\cite{deng2018image} and ECN~\cite{zhong2019invariance}, and comparable to UDAP~\cite{song2020unsupervised} on mAP, which proves the robustness of our method for unsupervised vehicle Re-ID.
%

%Furthermore, we observe that, 1) our method in target-only fashion (VAPC\_TO) achieves higher improvement on VeRi-776 than VeRi-Wild.
%
%We infer that VeRi-776 dataset, which contains five different viewpoints, is more suitable for our method to capture the viewpoint changes, comparing to VeRi-Wild dataset with mainly front and rear viewpoints. 
%
%In other words, our method is qualified for more complex scenarios with diverse viewpoint changes.
%
%2) Even the direct transfer fashion of our method (VAPC\_DT) achieves comparable performance as in domain adaption fashion (VAPC\_DA) on VeRi-Wild dataset.
%
%The main reason is the scenario of the source domain dataset VehicleID present similar as the tartget domain dataset VeRi-Wild, with both day and night data in mainly two viewpoints (front and rear).
%
%Therefore, the direct transfer of our method achieves great progress without any domain adaption scheme on VeRi-Wild while is overshadowed on VeRi-776 with more viewpoints.
%
%Anyway, our method in the domain adaption fashion (VAPC\_DA) makes up the robustness on VeRi-776 dataset.
%

Note that our method in target-only fashion (VAPC\_TO) even surpasses most unsupervised domain adaptation methods such as SPGAN~\cite{deng2018image} and ECN~\cite{zhong2019invariance}, and works comparably to UDAP~\cite{song2020unsupervised}.
This further verifies the promising performance of our method while handling unsupervised vehicle Re-ID especially without prior annotation information or source data.

\subsection{Ablation Study.}
In this section, we will thoroughly analyze the effectiveness of three critical components in the VAPC framework, including the two-period (tP) clustering strategy based on viewpoint prediction, $k$-reciprocal encoding (kR) and noise selection (NS), as reported in TABLE~\ref{abl}.
%
%
%Experimental results are reported in  TABLE~\ref{abl}, where the performance of remove all strategies and the model degenerates to only use the DBSCAN algorithm for global clustering is also reported as the baseline, see Table~\ref{abl} (e).
%
%We evaluate the effectiveness of our method by comparing with the baseline. In the case of known ground truth, we use Adjusted Mutual Information (AMI) to measure the clustering quality of our method and other methods. The larger the value of AMI, the more it matches the ground truth.
%In addition, in order to verify the importance of progressive clustering by viewpoint, we further analyze the two periods separately.
\begin{figure*}[t]
  \centering
  \includegraphics[height=7cm,width=18cm]{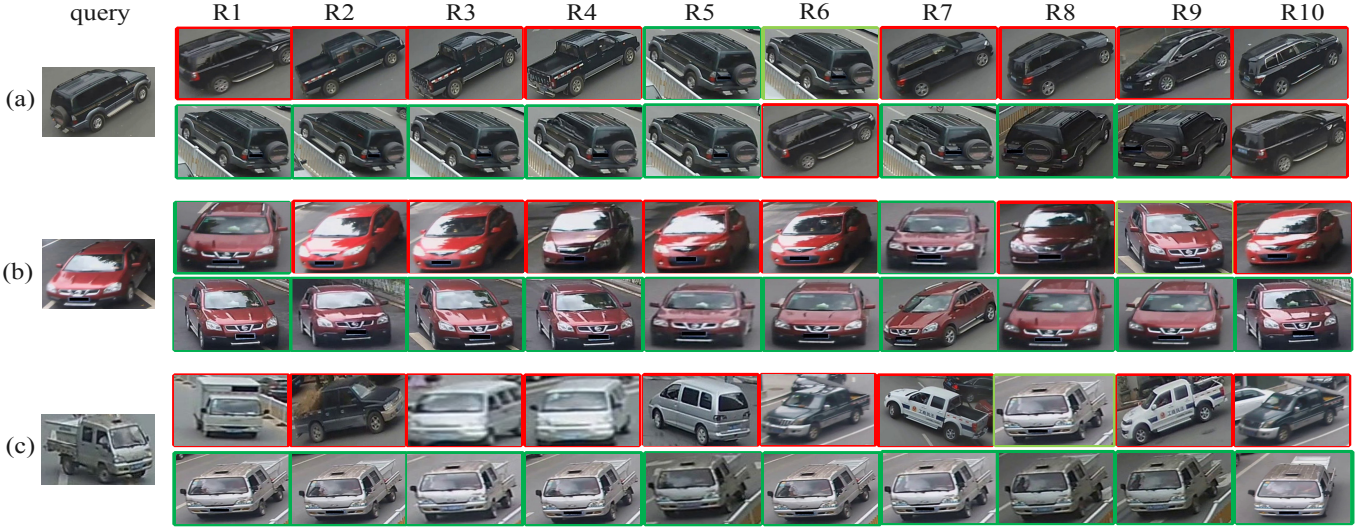}
  \caption{Examples of ranking results with and without noise selection on VeRi-776 dataset. For each query, the top and the bottom rows show the ranking result without and with noise selection, respectively. The green and red boxes indicate the right and the wrong matchings, respectively.}
  \label{xlsy-zs}
\vspace{-0.4cm}
\end{figure*}

\begin{figure}[t]
\centering
\includegraphics[height=6.5cm,width=8.8cm]{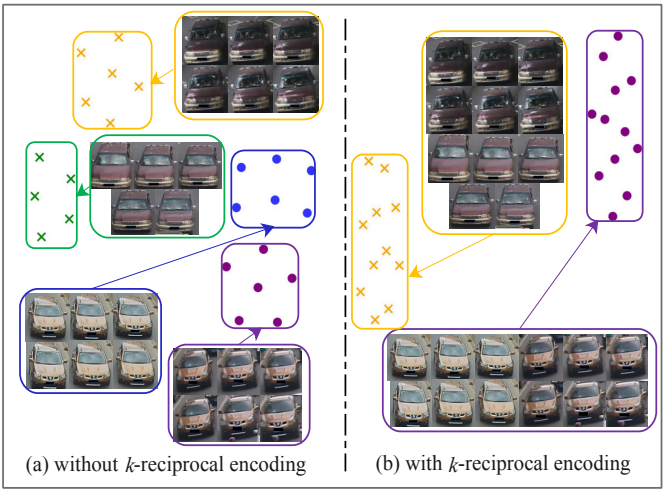}
%\caption{fig1}
\caption{Clustering illustrations with and without distance metric by $k$-reciprocal encoding on VeRi-776. The same shape represents the same identity, and the same color represents the same cluster.}
\label{k-coding}
\vspace{-0.55cm}
\end{figure} 

\noindent
{\bf Quantitative study.} One of the key contributions of our progressive clustering is the two-period clustering on both the same and different viewpoints for vehicle Re-ID. 
As shown in TABLE~\ref{abl} (b), without divide the viewpoints and removing the two-period (tP) strategy, we cluster all training samples directly after recognition stage, both mAP and rank scores significantly drop, -7.5\% in Rank-1 and -5.4\% in mAP on VeRi-776~\cite{liu2016deep}, while -3.6\% and -2.7\% on VeRi-Wild~\cite{lou2019large} test-3000, which verifies the effectiveness of the progressive clustering for unsupervised vehicle Re-ID.
Similar phenomenons happen to the $k$-reciprocal encoding (kR) and the noise selection (NS), as shown in TABLE~\ref{abl} (c) and TABLE~\ref{abl} (d). 
By removing the corresponding components, both mAP and rank scores significantly decline, which evidences the role of each component.
Without any of the three components, the baseline (as shown in TABLE~\ref{abl} (e)) results in stumble performance on both datasets due to the inability to cope with the various challenges brought about by the extreme viewpoint changes of vehicles.
By integrating all the three components, our method, as shown in TABLE~\ref{abl} (a) achieves promising results for unsupervised Re-ID.

\noindent
{\bf Qualitative study.}
To further understand the contribution of the three components, we visualize the results of different variants as discussed in Table~\ref{abl} in terms of sample distribution or ranking list, as shown in Fig.~\ref{fs} to Fig.~\ref{k-coding}.
From Fig.~\ref{fs} (a), we can see that more hard negative samples (different identities with highly similar appearance) with the same viewpoints tend to cluster without a two-period clustering strategy.
Our method successfully gathers vehicle images with diverse viewpoints, even with large appearance differences due to the viewpoint and illumination changes.
This further evidence the effectiveness of the proposed two-period clustering strategy, which can distinguish small gaps between different identities in the same viewpoint and mine the same identity samples with large gaps between different viewpoints.
%zheng-update
%
The role of $k$-reciprocal encoding is to mine samples sharing the most similar features despite appearance differences.
As shown in Fig~\ref{k-coding} (a), the result without $k$-reciprocal encoding tends to split the same identity with difference appearance caused by viewpoint and illumination changes into individual clusters, while it can merge them into one single cluster after introducing the $k$-reciprocal encoding, as shown in Fig~\ref{k-coding} (b).
Fig.~\ref{xlsy-zs} demonstrates the qualitative comparison of ranking results of three queries with or without noise selection. Clearly, after introducing the noise selection scheme, our method can hit more correct matchings in the earlier rankings and can remove the false matchings with a similar appearance as the queries.
\subsection{Analysis of Clustering Quality.}
Clustering quality is a crucial factor in clustering-based methods for vehicle Re-ID.
Therefore, we measure the clustering quality via
Adjusted Mutual Information (AMI)~\cite{vinh2010information} on our method compared to the state-of-the-art methods. 
AMI measures the distribution of ground truth and pseudo labels generated by clustering through mutual information.
The larger AMI, the closer distribution of the ground truth and pseudo labels, which in turn means the better clustering quality. 
We compare our method with Bottom~\cite{lin2019bottom}, k-means~\cite{kanungo2002efficient} and DBSCAN~\cite{ester1996density} which also allocate pseudo labels during clustering.

\begin{figure}[t]
\centering
\includegraphics[height=6.5cm,width=9cm]{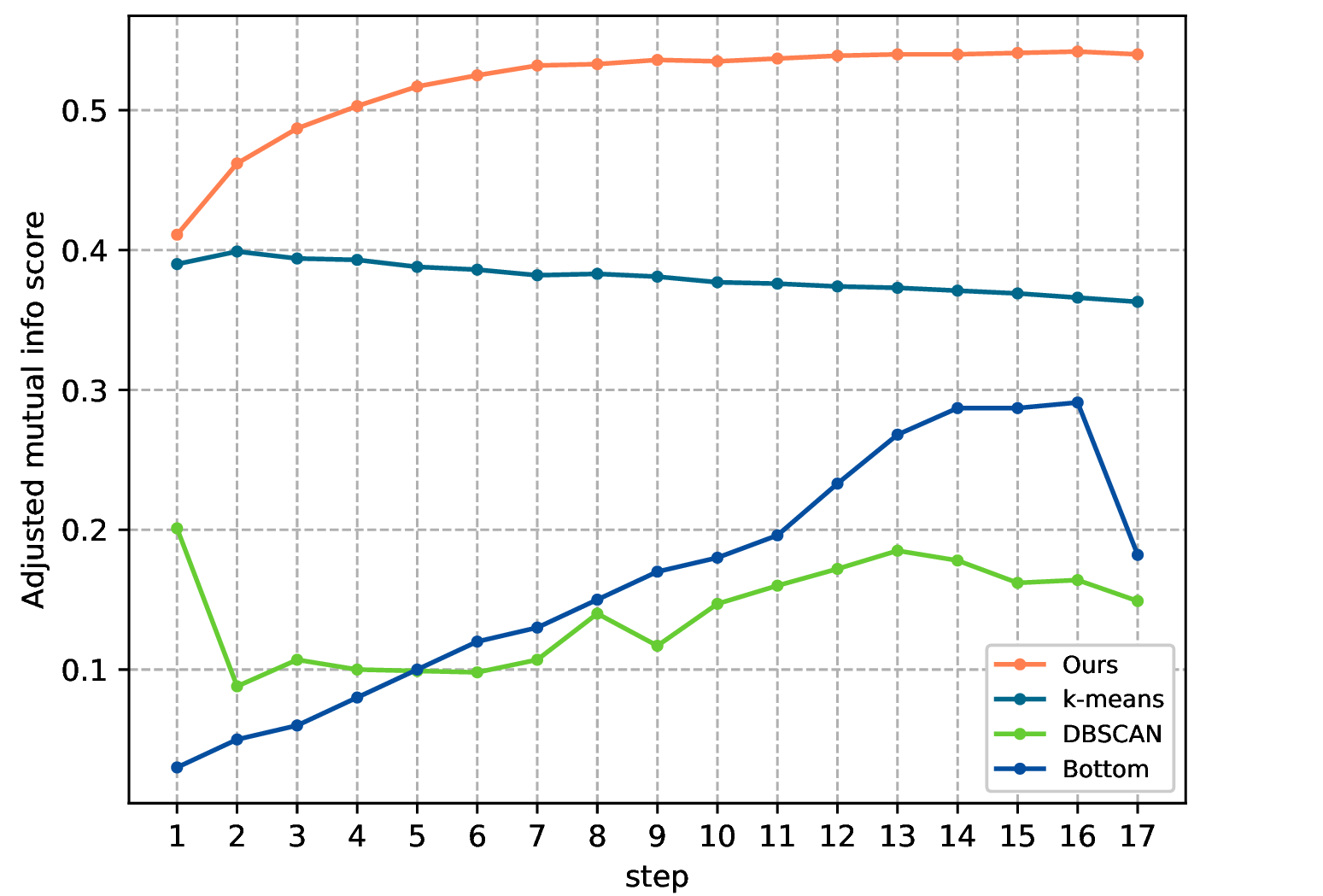}	
\caption{ The performance of clustering quality (AMI) on VeRi-776. Each step represents an iteration of progressive clustering and retraining the model.}
\label{clu}
\vspace{-0.3cm}
\end{figure}

As illustrated in Fig.~\ref{clu}, the classic clustering algorithm k-means~\cite{kanungo2002efficient} and DBSCAN~\cite{ester1996density} work stumblingly in the global comparison fashion.  
Furthermore, k-means~\cite{kanungo2002efficient} specifies the number of clusters, which makes the change of samples in the cluster relatively stable. However, due to global comparison, a large number of samples with the same viewpoint and different identities appear in the same cluster, which makes model training continue to decline.
%zheng-not convincing. why specify number result in performing smoothly? why global reusult in false instance and then slowly decline? think it carefully.
%
DBSCAN~\cite{ester1996density} is sensitive to noise; therefore, a large number of noise samples under various challenges in real scenes deteriorates the clustering quality.
Bottom~\cite{lin2019bottom} causes the final collapse due to the accumulation of the number of clustering errors each step.
Since clustering based on viewpoint division greatly simplifies the clustering task, and the strategy of progressively merges different viewpoints and gradually gathers vehicles of the same identity from different viewpoints, our method continues to improve with training.

\subsection{Investigation of Viewpoint Prediction.}
\begin{figure}[t]
\centering
\includegraphics[height=6.5cm,width=9cm]{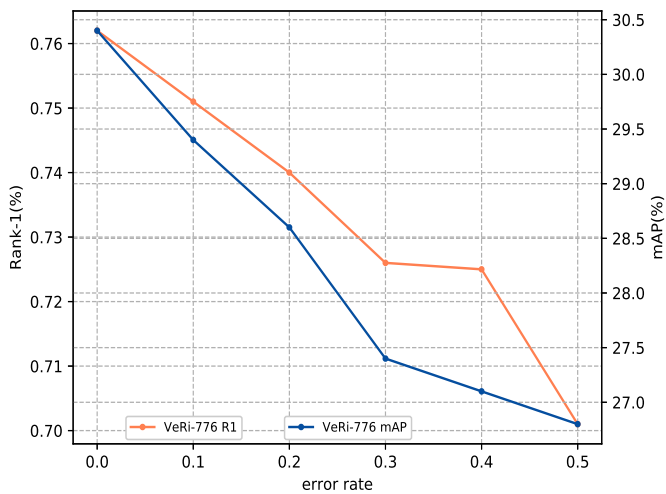}	
\caption{ The performance along with different error rate viewpoint predictors on VeRi-776.}
\label{view}
\vspace{-0.4cm}
\end{figure}

Viewpoint prediction is a prerequisite component in our framework, as discussed in~\ref{VP}.
To investigate the influence of viewpoint prediction in our method, we have trained a series of classifiers with different accuracy rates for viewpoint prediction. 
The experimental results are shown in Fig.~\ref{view}.
As expected, the Re-ID accuracy of VAPC\_TO decreases as the accuracy of the viewpoint classification classifier decreases. 
When the accuracy of the viewpoint classifier drops to 0.5, the accuracy of Rank-1 drops from 76.2\% to 70\%, (-6.2\%) on VeRi-776~\cite{liu2016deep}.
Even though our method with only 0.5 viewpoint classifier accuracy still outperforms the most unsupervised algorithms, as shown in TABLE~\ref{sota}. We can see that a robust viewpoint classifier can significantly improve the performance of our algorithm. And due to our more reasonable clustering strategy and effective noise processing, we can also perform well on a poor viewpoint classifier.

\begin{figure}
\centering
\subfigure[The parameter $ti$]{
\centering
\includegraphics[height=6.5cm,width=9cm]{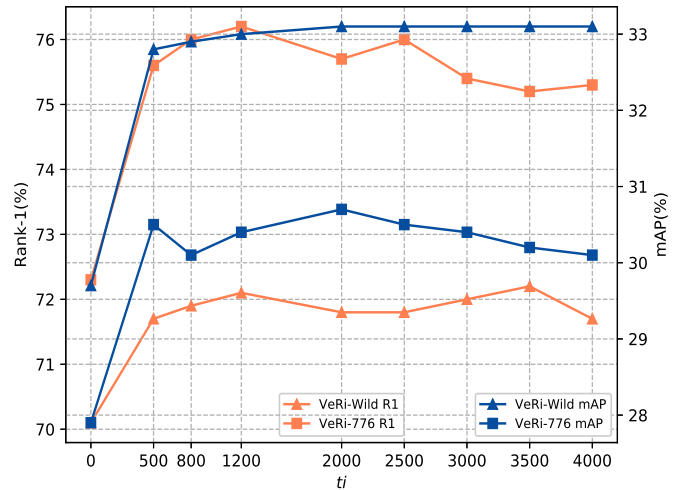}
%\caption{fig1}
}%

\subfigure[The parameter $\tilde{k}$]{
\centering
\includegraphics[height=6.5cm,width=9cm]{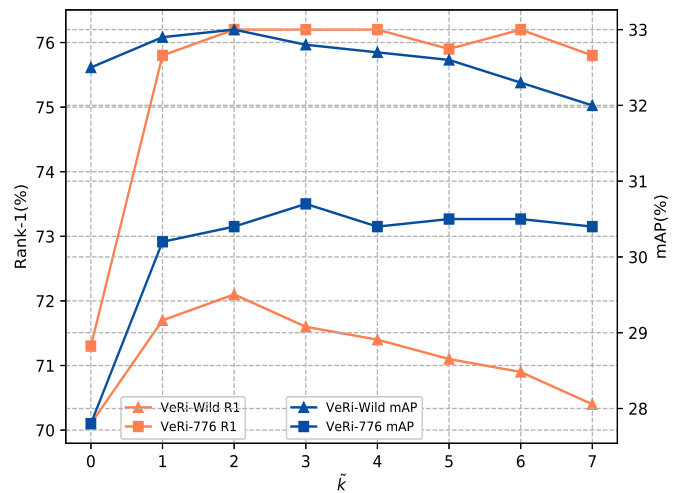}
%\caption{fig2}
}%
\caption{ Parameter and method analysis. (a) The impact of $ti$ in progressive clustering. (b) The impact of the number of $\tilde{k}$ in noise selection.}
\label{xlsy}
\vspace{-0.5cm}
\end{figure}
\subsection{ Parameter Analysis.}
\noindent
There are two essential parameters in our methods, $ti$ denoting the distance of the $ti$-th sample pair as the threshold for combining different viewpoint clusters as explained in~\ref{pc1} Eq~\eqref{R2}, and $\tilde{k}$ in Eq~\eqref{noise_sel}, indicating the judgment condition when selecting noise as explained in~\ref{pc1}.
We shall evaluate the impact of these two parameters in this section.

\noindent
\textbf{The impact of the number $ti$.}
As shown in Fig.~\ref{xlsy} (a), we vary $ti$ from 0 to 4000 to calculate the distance threshold $\tau$ and test the model performance.
$ti=0$ means only the same viewpoint clustering.
bigger $ti$, larger threshold $\tau$. 
A large $ti$ will harm the model performance. For example, when $ti>3500$ , a substantial performance drop can be observed. 
This is because the over large $ti$ may cause too many clusters of different viewpoints to be merged at one time, which resulting in a large number of incorrect classifications. 
However, over small $ti$ selects a few correct clusters, which also leads to poor performance. 
For the comprehensive performance of $ti$ on VeRi-776~\cite{liu2016deep} and VeRi-Wild~\cite{lou2019large}, we set $ti$ to 1200.

\noindent
\textbf{The impact of the number $\tilde{k}$.}
Fig.~\ref{xlsy} (b) reports the analysis on $\tilde{k}$ during the noise selection. As discussed in~\ref{pc1}, $\tilde{k}$ plays the role of limiting noise combined with clusters or other noises.
the larger $\tilde{k}$, the weaker limitation. 
The larger $\tilde{k}$ declines the performance on VeRi-Wild~\cite{lou2019large}, while remaining stable on VeRi-776~\cite{liu2016deep}. 
The reason is VeRi-Wild has a smaller inter-class difference compared to VeRi-776~\cite{liu2016deep}.
When $\tilde{k}$ increases, the constraint of judging whether the two clusters are merged is weakened, which increases the error rate. 
Based on the results on Fig.~\ref{xlsy} (b), we set $\tilde{k}=2$ for the best balance.

\section{Conclusion}
In this paper, we propose a viewpoint-aware progressive clustering method to solve the unsupervised Re-ID problem of vehicles. We analyzed the similarity dilemma of vehicle comparison, and for the first time explored the progressive clustering by dividing the training set into different subsets according to the viewpoint. In addition, we propose a noise selection strategy to solve the noise problem generated in the clustering process. Extensive experimental results demonstrate the effectiveness of the proposed methods in unsupervised Vehicle Re-ID.

Our method is based on the observation that images of vehicles from adjacent views normally share a large degree of common appearance, therefore they can be merged during clustering. 
However, it is still difficult to cluster the vehicles with only two viewpoints with large discrepancies, such as \textit{front} and \textit{rear}. In the future, we will further explore a more effective method to deal with these more challenging situations.

% if have a single appendix:
%\appendix[Proof of the Zonklar Equations]
% or
%\appendix  % for no appendix heading
% do not use \section anymore after \appendix, only \section*
% is possibly needed

% use appendices with more than one appendix
% then use \section to start each appendix
% you must declare a \section before using any
% \subsection or using \label (\appendices by itself
% starts a section numbered zero.)
%

%\appendices
%\section{Proof of the First Zonklar Equation}
%Appendix one text goes here.

% you can choose not to have a title for an appendix
% if you want by leaving the argument blank
%\section{}
%Appendix two text goes here.

% use section* for acknowledgment

% Can use something like this to put references on a page
% by themselves when using endfloat and the captionsoff option.
\ifCLASSOPTIONcaptionsoff
  \newpage
\fi

% trigger a \newpage just before the given references
% number - used to balance the columns on the last page
% adjust value as needed - may need to be readjusted if
% the document is modified later
%\IEEEtriggeratref{8}
% The "triggered" command can be changed if desired:
%\IEEEtriggercmd{\enlargethispage{-5in}}

% references section

% can use a bibliography generated by BibTeX as a .bbl file
% BibTeX documentation can be easily obtained at:
% http://mirror.ctan.org/biblio/bibtex/contrib/doc/
% The IEEEtran BibTeX style support page is at:
% http://www.michaelshell.org/tex/ieeetran/bibtex/
\bibliographystyle{IEEEtran}
% argument is your BibTeX string definitions and bibliography database(s)
\bibliography{Zheng_Viewpoint_aware_Progressive_Clustering_for_Unsupervised_Vehicle_Re-identification}
\end{document}